\documentclass{article}

\usepackage[final]{nips_2016}
\usepackage[utf8]{inputenc} 
\usepackage[T1]{fontenc}    
\usepackage{hyperref}       
\usepackage{url}            
\usepackage{booktabs}       
\usepackage{amsfonts}       
\usepackage{microtype}      

\usepackage{amssymb, amsmath}
\usepackage{epsfig}
\usepackage{array}
\usepackage{ifthen}
\usepackage{color}
\usepackage{fancyhdr}
\usepackage{graphicx}
\usepackage{mathtools}
\usepackage{csquotes}
\usepackage{xcolor}
\usepackage{multirow}
\newcommand\crule[3][black]{\textcolor{#1}{\rule{#2}{#3}}}

\newcommand{\E}{\textbf{E}}

\newcommand{\Cov}{\text{Cov}}
\newcommand{\Var}{\text{Var}}

\newcommand{\comm}[1]{}

\definecolor{color1}{RGB}{128,13,13}
\definecolor{color2}{RGB}{70,128,13}
\definecolor{color3}{RGB}{13,128,128}
\definecolor{color4}{RGB}{70,13,128}

\title{Estimating Mutual Information from Average Classification Error}

\author{
  Charles Y.~Zheng \\
  Department of Statistics\\
  Stanford University\\
  Stanford, CA 94305 \\
  \texttt{snarles@stanford.edu} \\
  \And
  Yuval ~Benjamini \\
  Department of Statistics \\
  Hebrew University\\
  Jerusalem, Israel\\
  \texttt{yuval.benjamini@mail.huji.ac.il}
}

\begin{document}

\maketitle

\begin{abstract}
Multivariate pattern analyses approaches in neuroimaging are
fundamentally concerned with investigating the quantity and type of
information processed by various regions of the human brain;
typically, estimates of classification accuracy are used to quantify
information.  While a extensive and powerful library of methods can be
applied to train and assess classifiers, it is not always clear how to
use the resulting measures of classification performance to draw
scientific conclusions: e.g. for the purpose of evaluating redundancy
between brain regions.  An additional confound for interpreting
classification performance is the dependence of the error rate on the
number and choice of distinct classes obtained for the classification
task.  In contrast, mutual information is a quantity defined independently of the experimental design, and has
ideal properties for comparative analyses.  
Unfortunately, estimating the mutual information based on observations
becomes statistically infeasible in high dimensions without some kind
of assumption or prior.
 
In this paper, we construct a novel classification-based estimator of mutual information
based on high-dimensional asymptotics. We show that in a particular limiting
regime, the mutual information is an invertible function of the
expected $k$-class Bayes error.  While the theory is based on a
large-sample, high-dimensional limit, we demonstrate through
simulations that our proposed estimator has superior
performance to the alternatives in problems of moderate
dimensionality. 
\end{abstract}

\section{Introduction}
A fundamental challenge of computational neuroscience is to understand
how information about the external world is processed and represented
in the brain. Each individual neuron aggregates the incoming
information into a single sequence of spikes--an output which is too
simplistic by itself to capture the full complexity of sensory
input. Only by combining the signals from massive ensembles of neurons
is it possible to reconstruct our complex representation of the
world. Nevertheless, neurons form hierarchies of specialization within
neural circuits, which are further organized in various specialized
regions of the brain.  At the lowest level of the hierarchy--individual neurons,
it is possible to infer and interpret the functional relationship between
a neuron and stimulus features of interest using single-cell recording technologies.
Due to the inherent stochasticity of the neural output, it is natural to
view the neuron as a noisy channel, and use mutual information
to quantify how much of the stimulus information is encoded by the neuron.
Moving up the hierarchy to the the macroscale level of organization in the
brain requires both different experimental methodologies and 
new approaches for summarizing and inferring measures of
information in the brain.

Shannon's mutual information $I(X; Y)$ is fundamentally a measure of dependence
between random variables $X$ and $Y$, and is defined as
\[
I(X;Y) = \int p(x, y) \log \frac{p(x, y)}{p(x)p(y)}dxdy.
\]
Various properties of $I(X; Y)$ make it ideal for quantifying the information between
a random stimulus $X$ and the signaling behavior of an ensembles of neurons, $Y$ [1].
A leading metaphor is that of a noisy communications channel; the mutual
information describes the rate at which $Y$ can communicate bits from
$X$.  This framework is well-suited for summarizing the properties of
a single neuron coding external stimulus information; indeed,
experiments studying the properties of a single or a small number of
neurons often make use of the concept of mutual information in
summarizing or interpreting their results [2]. See discussions in [3].
However, estimating mutual information for multiple channels requires large and
over-parameterized generative models.  

Machine learning algorithms showed a way forward: a seminal work
by Haxby [4] proposed to quantify the information in multiple
channels by measuring how well the stimulus can be identified from the
brain responses, in what is known as ``multivariate pattern analysis''
(MVPA). To demonstrate that a particular brain region responds to a
certain type of sensory information, one employs supervised learning
to build a classifier that classifies the stimulus class from the
brain activation in that region. Classifiers that achieve above-chance
classification accuracy indicate that information from the stimulus is
represented in the brain region. In principle, one could just as well
test the statistical hypothesis that the Fisher information or mutual
information between the stimulus and the activation patterns is
nonzero. But in practice, the machine learning approach enjoys several
advantages: First, it is invariant to the parametric representation of
the stimulus space, and is opportunistic in the parameterization of
the response space. This is an important quality for naturalistic
stimulus-spaces, such as faces or natural images. Second, it scales
better with the dimensionality of both the stimulus space and the
responses space, because a slimmer discriminative model can be used
rather than a fully generative model.

Nevertheless, classification error is problematic for quantifying the
strength of the relation between stimulus and outputs due to its
arbitrary scale and strong dependence on experimental
choices. Classification accuracy depends on the particular choice of
stimuli exemplars employed in the study and the number of partitions ($k$)
used to define the classes for the classification task. The difficulty
of the classification task depends on the number of classes defined:
high classification accuracy can be achieved relatively easily by
using a coarse partition of stimuli exemplars into classes. Often $k$ is an arbitrary design
constraint, and researchers try to extrapolate the error for alternative number of classes [5]. 
In a meta-analysis on visual decoding, Coutanche et al (2016) [6] quantified
the strength of a classification study using the formula
\[
\text{decoding strength} = \frac{\text{accuracy} - \text{chance}}{\text{chance}}.
\]
Such an approach may compensate for the differences in accuracy due
purely to choice of number of classes defined; however, no theory is
provided to justify the formula. In contrast, mutual information has
ideal properties for quantitatively comparing information between
different studies, or between different brain regions, subjects,
feature-spaces, or modalities. Not only is the mutual
information defined independently of the arbitrary definition of
stimulus classes (albeit still dependent on an implied distribution
over stimuli), it is even meaningful to discuss the difference between
the mutual information measured for one system and the mutual
information for a second system.

Hence, a popular approach which combines the strengths of the machine
learning approach and the advantages of the information theoretic
approach is to obtain a lower bound on the mutual information by using
the confusion matrix of a classifier.  Treves [7] first proposed using the empirical mutual information of
the classification matrix in order to obtain a lower bound of the
mutual information $I(X; Y)$; this confusion-matrix-based lower bound
has subsequently enjoyed widespread use in the MVPA literature
[2].  Even earlier that this, the idea of linking
classification performance to mutual information can be found in the
beginnings of information theory.
Fano's inequality provides a lower bound on mutual information in relation to
the optimal prediction error, or Bayes error. In practice, the bound obtained may be a vast
underestimate [8].

\subsection{Our contributions}

In this paper, we propose a new way to link classification performance
to the implied mutual information. To create this link we need to overcome the arbitrary choice of
exemplars, and the arbitrary number of classes k.  Towards this end,
we define a notion of $k$-class \emph{average Bayes error} which is
uniquely defined for any given stimulus distribution and stochastic
mapping from stimulus to response.  The $k$-class average Bayes error
is the expectation of the Bayes error (the classification error
of the optimal classifier) when $k$ stimuli exemplars are drawn
i.i.d. from the stimulus distribution, and treated as distinct
classes.  Hence the average Bayes error can in principle be estimated
if the appropriate randomization is employed for designing the
experiment.

Specifically, we establish
a relationship between the mutual information $I(X; Y)$ and the
average $k$-class Bayes error, $e_{ABE, k}$.  In short, we will
identify a function $\pi_k$ (which depends on $k$),
\begin{equation}\label{abepi}
e_{ABE, k} \approx \pi_k(\sqrt{2 I(X; Y)})
\end{equation}
and that this approximation becomes accurate under a limit where $I(X;
Y)$ is small relative to the dimensionality of $X$, and under the
condition that the components of $X$ are approximately independent.
The function $\pi_k$ is given by
\[
\pi_k(c) = 1 - \int_{\mathbb{R}} \phi(z - c)  \Phi(z)^{k-1} dz.
\]
This formula is not new to the information theory literature: it
appears as the error rate of an orthogonal constellation [9].  What
is surprising is that the same formula can be used to approximate the
error rate in much more general class of classification
problems\footnote{An intuitive explanation for this fact is that
  points from any high-dimensional distribution lie in an orthogonal
  configuration with high probability.}--this is precisely the
universality result which provides the basis for our proposed
estimator.

\begin{figure}
\centering
\begin{tabular}{ccrl}
\multirow{5}{*}{\includegraphics[scale = 0.5, clip=true, trim=0 0.2in 0 0.5in]{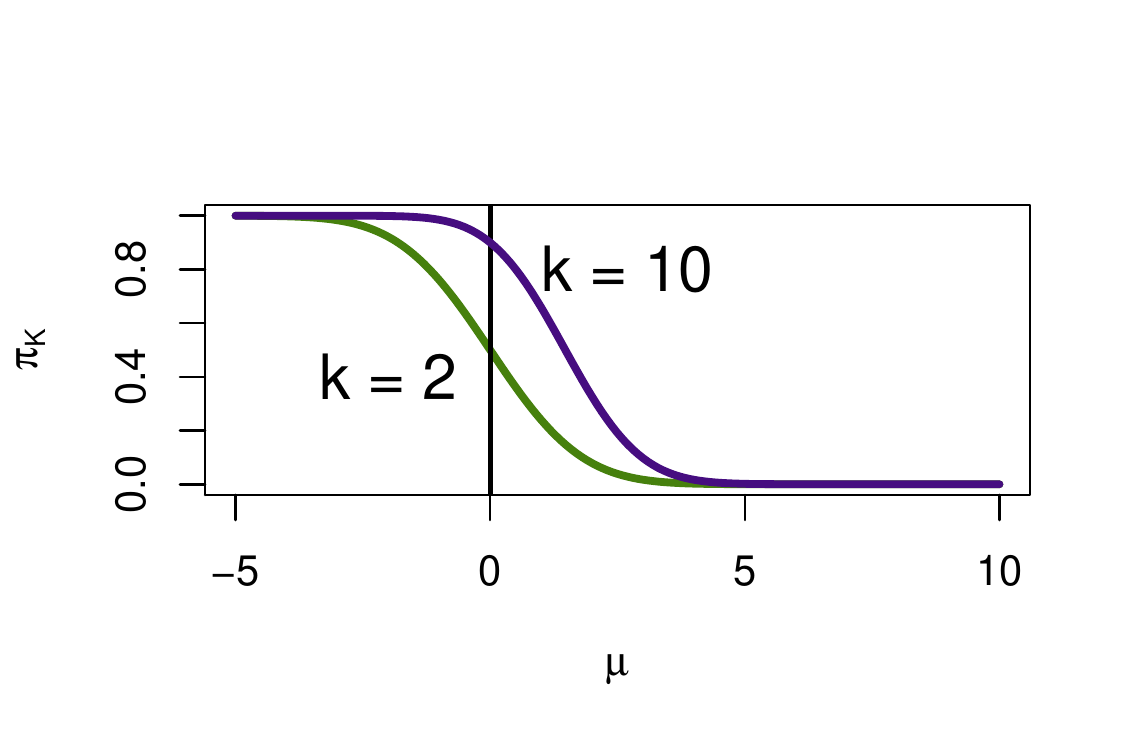}} &
\multirow{5}{*}{\includegraphics[scale = 0.5, clip=true, trim=0 0.2in 0.4in 0.5in]{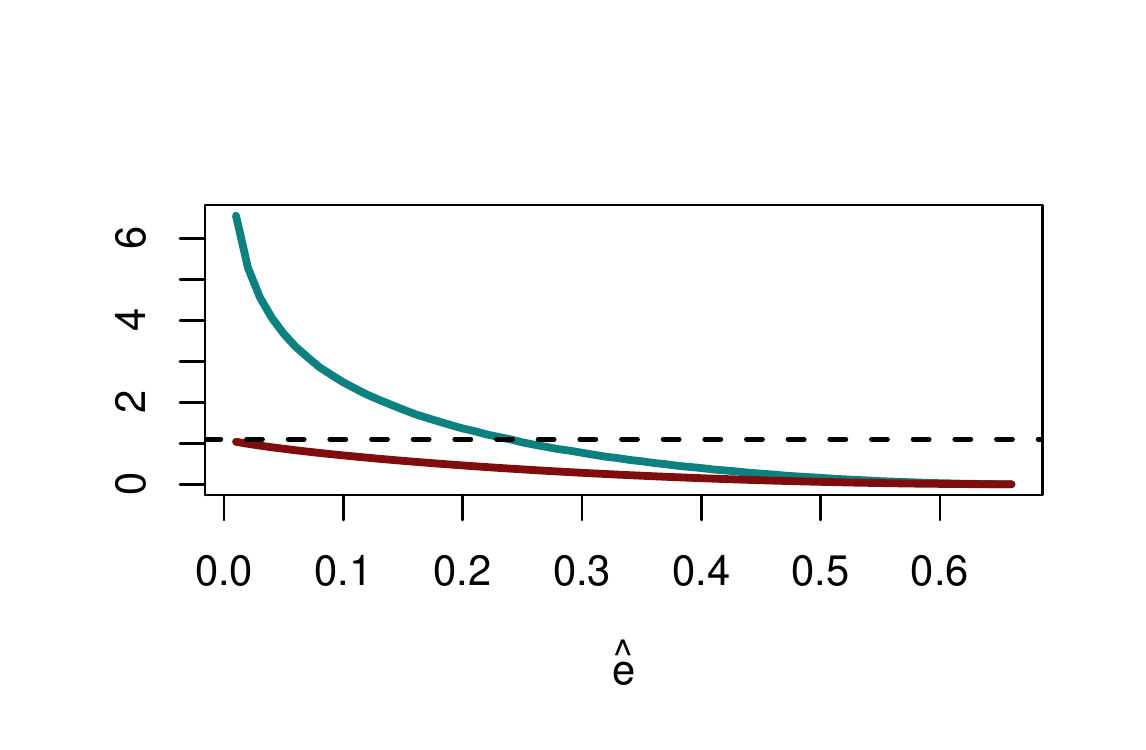}} & & \\
& & & \\
& & \crule[color3]{0.2cm}{0.2cm} & $\hat{I}_{HD}$\\
& & \crule[color1]{0.2cm}{0.2cm} & $\hat{I}_{Fano}$\\
& & & \\
& & & \\
& & & \\
& & & 
\end{tabular}
\caption{Left: The function $\pi_k(\mu)$ for $k = \{2, 10\}$.
Right: $\hat{I}_{HD}$ with $\hat{I}_{Fano}$ as functions of $\hat{e}_{gen}$, for $k = 3$.
While $\hat{I}_{Fano}$ is bounded from above by $\log(k)$ (dotted line), $\hat{I}_{HD}$ is unbounded.
\label{fig:pi}}
\end{figure}

Figure \ref{fig:pi} displays the plot of $\pi_k$ for several values of
$k$.  For all values of $k$, $\pi_k(\mu)$ is monotonically decreasing
in $\mu$, and tends to zero as $\mu \to \infty$, which is what we
expect since if $I(X; Y)$ is large, then the average Bayes error
should be small.  Another intuitive fact is that $ \pi_k(0) = 1 -
\frac{1}{k}, $ since after all, an uninformative response cannot lead
to above-chance classification accuracy.

The estimator we propose is
\[
\hat{I}_{HD} = \frac{1}{2}(\pi_{k}^{-1}(\hat{e}_{gen, \alpha}))^2,
\]
obtained by inverting the relation \eqref{abepi}, then substituting
an estimate of generalization error $\hat{e}_{gen, \alpha}$ for the $e_{ABE, k}$.  As such,
our estimator can be directly compared to the $\hat{I}_{Fano}$, since
both are functions of $\hat{e}_{gen,\alpha}$ (Figure 1.)
As the estimate of generalization error goes to zero, $\hat{I}_{Fano}$ approaches $\log(k)$
while $\hat{I}_{HD}$ goes to infinity.   This difference in behavior is due to the fact that
in contrast to Fano's inequality, the asymptotic relationship \eqref{abepi} is independent of the
number of classes $k$.

In the paper we argue for the advantages of our method in comparison to alternative discriminative estimators
under the assumption that the discriminative model approximates the Bayes rule.
While this is an unrealistic assumption, it simplifies the theoretical discussion, and allows us to clearly
discuss the principles behind our method. Alternatively, 
we can take the view that any observed classification error is a lower bound on the Bayes prediction, 
therefore interpreting our result as establishing a usually tighter lower bound on $I(X,Y)$.    

The organization of the paper is as follows.
We outline our framework in Section 2.1.
In Section 2.2 we present our key result, which links the asymptotic average Bayes error
to the mutual information, under an asymptotic setting intended to capture the
notion of high dimensionality\footnote{Namely, one where the number of classes
is fixed, and where the information $I(X; Y)$ remains fixed, while the
dimensionality of the input $X$ and output $Y$ both grow to infinity.
We make a number of additional regularity conditions to rule out
scenarios where $(X, Y)$ is really less ``high-dimensional'' than it
appears, since most of the variation is captured a low-dimensional
manifold.  }. In Section 2.3 we apply this result to
derive our proposed estimator, $\hat{I}_{HD}$ (where HD stands for
``high-dimensional.'')  Section 3 presents simulation results, and
Section 4 concludes.  All proofs are given in the supplement.

\section{Theory}

\subsection{Setting}

Let us assume that the variables $X, Y$ have a joint distribution $F$,
and that one can define a conditional distribution of $Y$ given $X$,
$Y|X \sim F_X,$ and let $G$ denote the marginal distribution of $X$.
We assume that data is collected using \emph{stratified sampling}.
For $j = 1,\hdots, k$, sample
  i.i.d. \emph{exemplars} $X^{(1)},\hdots, X^{(k)} \sim G$.  For $i =
  1,\hdots, n$, draw $Z^i$ iid from the uniform distribution on
  $1,\hdots, k$, then draw $Y^i$ from the conditional distribution
  $F_{X^{(Z_i)}}$.

Stratified sampling is commonly seen in controlled experiments, where an experimenter
chooses an input $X$ to feed into a black box, which outputs $Y$.  An
example from fMRI studies is an experimental design where the subject
is presented a stimulus $X$, and the experimenter measures the
subject's response via the brain activation $Y$. \footnote{Note the
  asymmetry in our definition of stratified sampling: our convention
  is to take $X$ to be the variable preceding $Y$ in causal order.
  Such causal directionality constrains the stratified sampling to
  have repeated $X$ rather than repeated $Y$ values, but has no
  consequence for the mutual information $I(X; Y)$, which is a
  symmetric function.}

When stratified sampling is employed, one can
define an \emph{exemplar-based} classification task.
One defines the \emph{class function}
$Z$ by
\[
Z: \{X^{(1)}, \hdots, X^{(k)}\} \to \{1,\hdots, k\},
\]
\[
Z(X^{(i)}) = i\text{ for }i = 1, \hdots, k.
\]
One defines the generalization error by
\begin{equation}\label{egendef}
e_{gen}(f) = \frac{1}{k} \sum_{i=1}^k\Pr[f(Y) \neq Z|X = X^{(i)}].
\end{equation}
In an exemplar-based classification, there is no need to
specify an arbitrary partition on the input space (as is the case in category-based classification),
but note that the $k$ classes are \emph{randomly} defined.  One consequence is that
the Bayes error $e_{Bayes}$ is a random variable: when the sampling
produces $k$ similar exemplars, $e_{Bayes}$ will be higher, and when
the sampling produces well-separated exemplars $e_{Bayes}$ may be
lower.  Therefore, it is useful to consider
the \emph{average Bayes error},
\begin{equation}\label{abedef}
e_{ABE, k} = \E_{X^{(1)},\hdots, X^{(k)}}[e_{Bayes}],
\end{equation}
where the expectation is taken over the joint distribution of
$X^{(1)},\hdots, X^{(k)} \stackrel{iid}{\sim} G$.

We use the terminology
\emph{classifier} to refer to any algorithm which takes data as input,
and produces a classification rule $f$ as output.  Mathematically
speaking, the classifier is a functional which maps a set of
observations to a classification rule, $ \mathcal{F}:
\{(x^{1},y^{1}),\hdots, (x^{m}, y^{m})\} \mapsto f(\cdot).  $ The data
$(x^1,y^1),\hdots, (x^m, y^m)$ used to obtain the classification rule
is called \emph{training data.}  When the goal is to obtain
\emph{inference} about the generalization error $e_{gen}$ of the
classification rule $f$, it becomes necessary to split the data into
two independent sets: one set to train the classifier, and one to
evaluate the performance.  The reason that such a splitting is
necessary is because using the same data to test and train a
classifier introduces significant bias into the empirical
classification error [10].  The classification
rule is obtained via $ f = \mathcal{F}(S_{train}), $ where $S_{train}$ is the training set,
 and the performance of the classifier is evaluated by predicting the classes
of the test set.  The results of this test are summarized by a $k
\times k$ \emph{confusion matrix} $M$ with $ M_{ij} = \sum_{\ell=r_1 +
  1}^r I(f(y^{(i), r}) = j).  $ The $i, j$th entry of $M$ counts how
many times a output in the $i$th class was classified to the $j$th
class.  The \emph{test error} is the proportion of off-diagonal terms
of $M$, $ e_{test} = \frac{1}{kr} \sum_{i \neq j} M_{ij}, $ and is an
unbiased estimator of $e_{gen}$.  However, in small sampling regimes
the quantity $e_{test}$ may be too variable to use as an estimator of
$e_{gen}$.  We recommend the use of Bayesian smoothing, defining an
$\alpha$-smoothed estimate $\hat{e}_{gen, \alpha}$ by $ \hat{e}_{gen,
  \alpha} = (1 - \alpha) e_{test} + \alpha \frac{k-1}{k}, $ which
takes a weighted average of the unbiased estimate $e_{test}$, and the
natural prior of \emph{chance classification}.

We define a discriminative estimator to be a function which maps the
misclassification matrix to a positive number, $ \hat{I}:
\mathbb{N}^{k \times k} \to \mathbb{R}.  $ We are aware of the
following examples of discriminative estimators: (1) estimators $\hat{I}_{Fano}$
derived from using Fano's inequality, and (2) the empirical
information of the confusion matrix, $\hat{I}_{CM}$, as introduced by Treves [7].
We discuss these estimators in Section 3.

\subsection{Universality result}

We obtain the universality result in two steps.  First, we link the
average Bayes error to the moments of some statistics $Z_i$.
Secondly, we use taylor approximation in order to express $I(X; Y)$ in
terms of the moments of $Z_i$.  Connecting these two pieces yields the
formula \eqref{abepi}.

Let us start by rewriting the average Bayes error:
\[
e_{ABE, k} = \Pr[p(Y|X_1) \leq \max_{j \neq 1} p(Y|X_j)| X = X_1].
\]
Defining the statistic $Z_i = \log p(Y|X_i) - \log p(Y|X_1)$, where $Y
\sim p(y|X_1)$, we obtain $ e_{ABE} = \Pr[\max_{j > 1} Z_i > 0].  $
The key assumption we need is that $Z_2,\hdots, Z_k$ are
asymptotically multivariate normal.  If so, the following lemma allows
us to obtain a formula for the misclassification rate.

\textbf{Lemma 1. }
\emph{
Suppose $(Z_1, Z_2, \hdots, Z_k)$ are jointly multivariate normal, with 
$\E[Z_1 - Z_i]= \alpha$, 
$\Var(Z_1) = \beta \geq 0$, 
$\Cov(Z_1, Z_i) = \gamma$, 
$\Var(Z_i)= \delta$, and $\Cov(Z_i, Z_j) = \epsilon$ for all $i, j = 2, \hdots,
k$, such that $\beta + \epsilon - 2\gamma > 0$.  Then, letting
\[
\mu = \frac{\E[Z_1 - Z_i]}{\sqrt{\frac{1}{2}\Var(Z_i - Z_j)}} = \frac{\alpha}{\sqrt{\delta - \epsilon}},
\]
\[
\nu^2 = \frac{\Cov(Z_1 -Z_i, Z_1 - Z_j)}{\frac{1}{2}\Var(Z_i - Z_j)} = \frac{\beta + \epsilon - 2\gamma}{\delta - \epsilon},
\]
we have
\begin{align*}
\Pr[Z_1 < \max_{i=2}^k Z_i] &= \Pr[W < M_{k-1}]
\\&= 1 - \int \frac{1}{\sqrt{2\pi\nu^2}} e^{-\frac{(w-\mu)^2}{2\nu^2}} \Phi(w)^{k-1} dw,
\end{align*}
where $W \sim N(\mu, \nu^2)$ and $M_{k-1}$ is the maximum of $k-1$
independent standard normal variates, which are independent of $W$.
}

To see why the assumption that $Z_2,\hdots, Z_k$ are multivariate normal might be justified, suppose that $X$ and $Y$ have the same dimensionality $d$, and that
joint density factorizes as
\[
p(x^{(j)}, y) = \prod_{i=1}^d p_i(x^{(j)}_i, y_i)
\]
where $x_i^{(j)}, y_i$ are the $i$th scalar components of the vectors $x^{(j)}$ and $y$.
Then,
\[
Z_i = \sum_{m=1}^d \log p_m(y_m | x^{(i)}_m) - \log p_m(y_m | x^{(m)}_1)
\]
where $x_{i, j}$ is the $i$th component of $x_j$.  The $d$ terms $\log
p_m(y_m | x_{m, i}) - \log p_m(y_m | x_{m, 1})$ are independent across
the indices $m$, but dependent between the $i = 1,\hdots, k$.
Therefore, the multivariate central limit theorem can be applied to
conclude that the vector $(Z_2,\hdots, Z_k)$ can be scaled to converge
to a multivariate normal distribution.  While the componentwise
independence condition is not a realistic assumption, the key property
of multivariate normality of $(Z_2,\hdots, Z_k)$ holds under more
general conditions, and appears reasonable in practice.

It remains to link the moments of $Z_i$ to $I(X;Y)$.  This is accomplished by approximating the logarithmic term by the Taylor expansion
\[
\log \frac{p(x, y)}{p(x) p(y)} \approx \frac{p(x, y) - p(x) p(y)}{p(x) p(y)} - \left(\frac{p(x, y) - p(x) p(y)}{p(x) p(y)}\right)^2 + \hdots.
\]
A number of assumptions are needed to ensure that needed
approximations are sufficiently accurate; and additionally, in order
to apply the central limit theorem, we need to consider a
\emph{limiting sequence} of problems with increasing dimensionality.
We now state the theorem.

\textbf{Theorem 1.} \emph{Let $p^{[d]}(x, y)$ be a sequence of joint densities
for $d = 1,2,\hdots$.  Further assume that
\begin{itemize}
\item[A1.] $\lim_{d \to \infty} I(X^{[d]}; Y^{[d]}) = \iota < \infty.$
\item[A2.] There exists a sequence of scaling constants $a_{ij}^{[d]}$
and $b_{ij}^{[d]}$ such that the random vector $(a_{ij}\ell_{ij}^{[d]} +
b_{ij}^{[d]})_{i, j = 1,\hdots, k}$ converges in distribution to a
multivariate normal distribution,
where $\ell_{ij} = \log p(y^{(i)}|x^{(i)})$ for independent $y^{(i)} \sim p(y|x^{(i)})$.
\item[A3.] Define \[
u^{[d]}(x, y) = \log p^{[d]}(x, y) - \log p^{[d]}(x) - \log p^{[d]}(y).
\]
There exists a sequence of scaling constants $a^{[d]}$, $b^{[d]}$ such that
\[
a^{[d]}u^{[d]}(X^{(1)}, Y^{(2)}) + b^{[d]}
\]
converges in distribution to a univariate normal distribution.
\item[A4.] For all $i \neq k$,
\[\lim_{d \to \infty}\Cov[u^{[d]}(X^{(i)}, Y^{(j)}), u^{[d]}(X^{(k)}, Y^{(j)})] = 0.\]
\end{itemize}
Then for $e_{ABE, k}$ as defined above, we have
\[
\lim_{d \to \infty} e_{ABE, k} = \pi_k(\sqrt{2 \iota})
\]
where
\[
\pi_k(c) = 1 - \int_{\mathbb{R}} \phi(z - c)  \Phi(z)^{k-1} dz
\]
where $\phi$ and $\Phi$ are the standard normal density function and
cumulative distribution function, respectively.}

Assumptions A1-A4 are satisfied in a variety of natural models.  One
example is a multivariate Gaussian sequence model where $X \sim N(0,
\Sigma_d)$ and $ Y = X + E $ with $ E \sim N(0, \Sigma_e), $ where
$\Sigma_d$ and $\Sigma_e$ are $d \times d$ covariance matrices, and
where $X$ and $E$ are independent.  Then, if $d \Sigma_d$ and
$\Sigma_e$ have limiting spectra $H$ and $G$ respectively, the joint
densities $p(x, y)$ for $d = 1,\hdots, $ satisfy assumptions A1 - A4.
Another example is the multivariate logistic model, which we describe
in Section 3.  We further discuss the rationale behind A1-A4 in the
supplement, along with the detailed proof.

\subsection{High-dimensional estimator}

As stated in the introduction, we propose the estimator
\[
\hat{I}_{HD}(M) = \frac{1}{2}(\pi_{k}^{-1}(\hat{e}_{gen, \alpha}))^2.
\]
For sufficiently high-dimensional problems, $\hat{I}_{HD}$ can
accurately recover $I(X; Y) > \log k$, supposing also that the
classifier $\mathcal{F}$ consistently estimates the Bayes rule.  The
number of observations needed depends on the convergence rate of
$\mathcal{F}$ and also the complexity of estimating $e_{gen, \alpha}$.
Therefore, without making assumptions on $\mathcal{F}$, the sample
complexity is at least exponential in $I(X; Y)$.  This is because when
$I(X; Y)$ is large relative to $\log(k)$, the Bayes error $e_{ABE, k}$
is exponentially small.  Hence $O(1/e_{ABE, k})$ observations in the
test set are needed to recover $e_{ABE, k}$ to sufficient precision.
While the sample complexity exponential in $I(X; Y)$ is by no means
ideal, by comparison, the nonparametric estimation approaches have a
complexity exponential in the dimensionality.  Hence, $\hat{I}_{HD}$
is favored over nonparametric approaches in settings with high
dimensionality and low signal-to-noise ratio.

\section{Simulation}

We compare the discriminative estimators $\hat{I}_{CM}$,
$\hat{I}_{Fano}$, $\hat{I}_{HD}$ with a nonparametric estimator
$\hat{I}_0$ in the following simulation, and the correctly specified parametric estimator $\hat{I}_{MLE}$.  We generate data according
to a multiple-response logistic regression model, where $ X \sim N(0,
I_p) $, and $Y$ is a binary vector with conditional distribution
\[
Y_i|X = x \sim \text{Bernoulli}(x^T B_i)
\]
where $B$ is a $p \times q$ matrix.  One application of this model
might be modeling neural spike count data $Y$ arising in response to
environmental stimuli $X$ [12].  We choose the naive Bayes for the
classifier $\mathcal{F}$: it is consistent for estimating the Bayes
rule. 

\begin{figure}
\begin{center}
\begin{tabular}{ccrl}
&\multirow{10}{*}{\includegraphics[scale = 0.5, clip=true, trim=0.4in 0.5in 0 0.5 in]{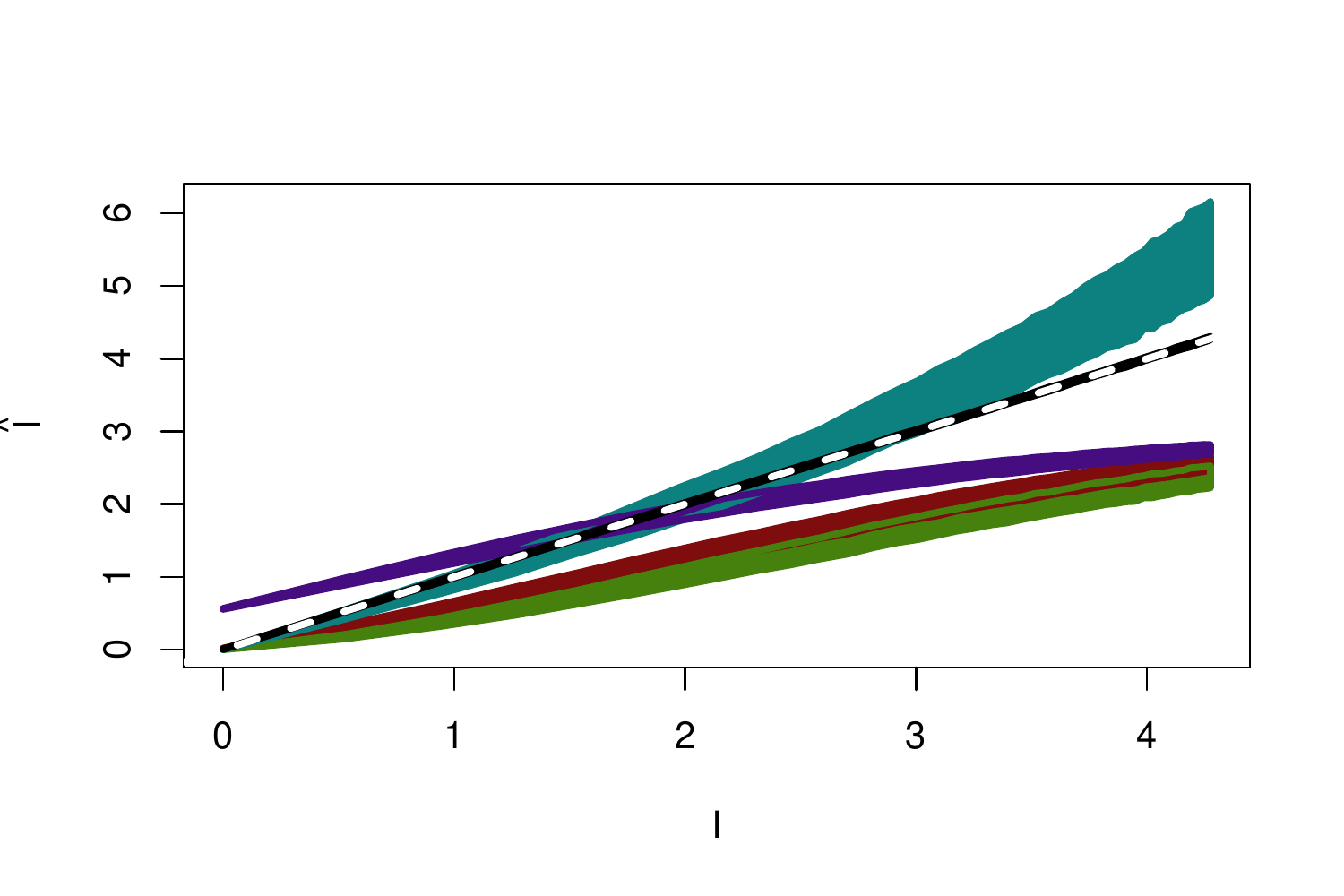}}&&\\
&  Sampling distribution of $\hat{I}$& &\\
& & &\\
& & \crule[color3]{0.2cm}{0.2cm} &$\hat{I}_{HD}$\\
& & \crule[black]{0.2cm}{0.2cm} &$\hat{I}_{MLE}$\\
 & &\crule[color4]{0.2cm}{0.2cm} &$\hat{I}_{0}$ \\
$\hat{I}$& & \crule[color1]{0.2cm}{0.2cm} &$\hat{I}_{CM}$\\
& & \crule[color2]{0.2cm}{0.2cm} &$\hat{I}_{Fano}$\\
& & &\\
& & &\\
& & &\\
& $I(X; Y)$& &\\
\end{tabular}
\end{center}
\caption{Sampling distributions of $\hat{I}$ for data generated from the multiple-response logistic model.  $p = q = 10$; $k= 20$; $B = sI_{10}$, where $s \in [0, \sqrt{200}]$; and  $r = 1000$.}
\end{figure}

The estimator $\hat{I}_{Fano}$ is based on Fano's inequality, which reads
\[
H(Z|Y) \leq H(e_{Bayes}) + e_{Bayes} \log ||\mathcal{Z}| - 1|
\]
where $H(e)$ is the entropy of a Bernoulli random variable with
probability $e$.  Replacing $H(Z|Y)$ with $H(X|Y)$ and replacing
$e_{Bayes}$ with $\hat{e}_{gen, \alpha}$, we get the estimator
\[
\hat{I}_{Fano}(M) = log(K) - \hat{e}_{gen, \alpha} log(K-1) + \hat{e}_{gen, \alpha} log(p) + (1-\hat{e}_{gen, \alpha}) log(1-\hat{e}_{gen, \alpha}).
\]
Meanwhile, the confusion matrix estimator computes
\[
\hat{I}_{CM}(M) = \frac{1}{k^2} \sum_{i=1}^k \sum_{j=1}^k \log \frac{M_{ij}}{r/k},
\]
which is the empirical mutual information of the discrete joint
distribution $(Z, f(Y))$.

It is known that $\hat{I}_{CM}$, $\hat{I}_0$ tend to underestimate the mutual information.
Quiroga et al. [2] discussed two sources of `information loss' which lead to $\hat{I}_{CM}$ underestimating the mutual information:
the discretization of the classes, and the error in approximating the Bayes rule.
Meanwhile, Gastpar et
al. [11] showed that $\hat{I}_0$ is biased downwards due to undersampling
of the exemplars: to counteract this bias, they introduce the
anthropic correction estimator $\hat{I}_\alpha$\footnote{However, without a
principled approach to choose the parameter $\alpha \in (0,1]$,
  $\hat{I}_\alpha$ could still vastly underestimate or overestimate
  the mutual information.}.

In addition to the sources of information loss discussed by Quiroga et al., an additional reason why $\hat{I}_{CM}$ and $\hat{I}_{Fano}$ underestimate the mutual information is that they are upper bounded by $\log(k)$, where $k$ is the number of classes.  As $I(X; Y)$ exceeds $\log(k)$,
the estimate $\hat{I}$ can no longer approximate $I(X; Y)$, even up to
a constant factor.  In contrast, $\hat{I}_{HD}$ is unbounded and 
may either underestimate or overestimate the mutual information in general,
but performs well when the high-dimensionality assumption is met.

In Figure 2 we show the sampling distributions of the five
estimators as $I(X; Y)$ is varied in the interval $[0, 4]$.  The estimator $\hat{I}_{MLE}$
is a plug-in estimator using $\hat{B}$, the coefficient matrix estimated via multinomial regression of $Y$ on $X$;
it recovers the true mutual information within $\pm 2\%$ with a probability of 90\%.
We see that $\hat{I}_{CM}$, $\hat{I}_{Fano}$, and $\hat{I}_0$ indeed begin to
asymptote as they approach $\log(k) = 2.995$.  In contrast,
$\hat{I}_{HD}$ remains a good approximation of $I(X; Y)$ within the
range, although it begins to overestimate at the right endpoint.  
The reason why $\hat{I}_{HD}$ loses accuracy as the true information $I(X; Y)$ increases is that
the multivariate normality approximation used to derive the estimator becomes less accurate when the conditional distribution 
$p(y|x)$ becomes highly concentrated.

\section{Discussion}

Discriminative estimators of mutual information have the potential to
estimate mutual information in high-dimensional data without resorting
to fully parametric assumptions.  However, a number of practical
considerations also limit their usage.  First, one has to find a good
classifier $\mathcal{F}$ for the data: techniques for model selection
can be used to choose $\mathcal{F}$ from a large library of methods.
However, there is no way to guarantee how well the chosen classifier
approximates the optimal classification rule.  Secondly, one has to
estimate the generalization error from test data: the complexity of
estimating $e_{gen}$ could become the bottleneck when $e_{gen}$ is
close to 0.  Thirdly, for previous estimators $\hat{I}_{Fano}$ and
$\hat{I}_{CM}$, the ability of the estimator to distinguish high
values of $I(X; Y)$ is limited by the number of classes $k$.  
Our estimator $\hat{I}_{HD}$ is subject to the first two limitations,
along with any conceivable discriminative estimator, but overcomes the
third limitation under the assumption of stratified sampling and high
dimensionality.

It can be seen that additional assumptions are indeed needed to overcome the third limitation,
the  $\log(k)$ upper bound.  Consider the following worst-case example: let $X$ and $Y$ have joint density $
p(x, y) = \frac{1}{k}I(\lfloor kx \rfloor = \lfloor ky \rfloor) $ on
the unit square.  Under partition-based classification, if we set
$Z(x) = \lfloor kx \rfloor + 1$, then no errors are made under the
Bayes rule. We therefore have a joint distribution which maximizes any
reasonable discriminative estimator but has \emph{finite} information
$I(X; Y) = \log(k)$.  The consequence of this is that under
partition-based classification, we cannot hope to distinguish
distributions with $I(X; Y) > \log(k)$.  The situation is more
promising if we specialize to stratified sampling: in the same
example, a Bayes of zero is no longer likely due to the possibility of
exemplars being sampled from the same bin (`collisions')--we obtain an
approximation to the average Bayes error through a Poisson sampling
model: $e_{ABE, k} \approx \frac{1}{e}\sum_{j=1}^\infty
\frac{1}{j(j!)}= 0.484$.  By specializing further to the high-dimensional regime,
we obtain even tighter control on the relation between Bayes error and mutual information.
Our estimator therefore provides more accurate estimation at the cost of more additional assumptions,
but just how restrictive are these assumptions?

The assumption of stratified sampling is usually not met in the most common applications of classification
where the classes are defined \emph{a priori}.  For instance, if the classes consist of three different species of iris,
it does not seem appropriate to model the three species as i.i.d. draws from some distribution on a space of infinitely many
potential iris species.  Yet, when the classes have been pre-defined in an arbitrary manner,
the mutual information between a latent class-defining variable $X$ and $Y$ may be only weakly related to the classification
accuracy.  We rely on the stratified sampling assumption to obtain the necessary control on how the classes
in the classification task are defined.  Fortunately, in many applications where one is interested in estimating $I(X; Y)$,
a stratified sampling design can be practically implemented.

The assumption of high dimensionality is not easy to check:
having a high-dimension response $Y$ does not suffice, since even then $Y$ could still
lie close to a low-dimensional manifold.  In such cases, $\hat{I}_{HD}$ could either overestimate or underestimate the mutual information.
In situations where $(X, Y)$ lie on a manifold, one
  could effectively estimate mutual information by would be to
  combining dimensionality reduction with nonparametric information
  estimation [13]. 
  We suggest the following diagnostic to determine if our method is appropriate: subsample within the classes collected
and check that $\hat{I}_{HD}$ does not systematically increase or decrease with the number of classes $k$.

The assumption of approximating the Bayes rule is impractical to check, 
as any nonparametric estimate of the Bayes error requires exponentially many observations.
Hence, while the present paper studies the `best-case' scenario where the model is well-specified,
it is even more important to understand the robustness of our method in the more realistic case
where the model is misspecified.  We leave this question to future work.

Even given a classifier which consistently estimates the Bayes error,
the estimator $\hat{I}_{HD}$ can still be improved.  One can employ more
sophisticated methods to estimate $e_{ABE, k}$: for example,
extrapolating from learning curves [14].  Furthermore,
depending on the risk function, one may debias or shrink the estimate
$\hat{I}_{HD}$ to achieve a more favorable bias-variance tradeoff.

All of the necessary assumptions are met in our simulation experiment, 
hence our proposed estimator is seen to dramatically outperform existing estimators. It remains to assess the utility
of our estimation procedure in a real-world example, where both the high-dimensional assumption and the model specification assumption are likely to be violated.  In a forthcoming
work, we apply our framework to evaluate visual encoding models in
human fMRI data.

\subsubsection*{Acknowledgments}

We thank John Duchi, Youngsuk Park, Qingyun Sun, Jonathan Taylor, Trevor Hastie, Robert Tibshirani for useful discussion.  CZ is supported by an NSF graduate research fellowship.

\section*{References}

\small

[1] Borst, A. \& Theunissen, F. E. (1999). ``Information theory and neural coding''
\emph{Nature Neurosci.}, vol. 2, pp. 947-957.

[2] Quiroga, R. Q., \& Panzeri, S. (2009). ``Extracting information from neuronal populations: information theory and decoding approaches''. \emph{Nature Reviews Neuroscience}, 10(3), 173-185.

[3] Paninski L. , ``Estimation of entropy and mutual information,'' \emph{Neural Comput.}, vol. 15, no. 6, pp. 1191-1253, 2003.

[4] Haxby, J.V., et al.  (2001). "Distributed and overlapping representations of faces and objects in ventral temporal cortex." \emph{Science} 293.5539: 2425-2430.

[5] Kay, K. N., et al. ``Identifying natural images from human brain activity.'' \emph{Nature} 452.7185 (2008): 352-355.

[6] Coutanche, M.N., Solomon, S.H., and Thompson-Schill S. L., ``A meta-analysis of fMRI decoding: Quantifying influences on human visual population codes.'' 
\emph{Neuropsychologia} 82 (2016): 134-141.

[7] Treves, A. (1997). ``On the perceptual structure of face space.'' \emph{Bio Systems}, 40(1-2), 189?96. 

[8] Beirlant, J., Dudewicz, E. J., Gy\:{o}rfi, L., \& der Meulen,
E. C. (1997). ``Nonparametric Entropy Estimation: An
Overview.'' \emph{International Journal of Mathematical and Statistical
Sciences}, 6, 17-40.

[9] Tse, D., \& Viswanath, P. (2005). \emph{Fundamentals of wireless
communication.} Cambridge university press, 

[10] Friedman, J., Hastie, T., \& Tibshirani, R. (2008). \emph{The elements
of statistical learning.} Vol. 1. Springer, Berlin: Springer series in
statistics.

[11] Gastpar, M.  Gill, P.  Huth, A. \& Theunissen, F. (2010). ``Anthropic
Correction of Information Estimates and Its Application to Neural
Coding.'' \emph{IEEE Trans. Info. Theory}, Vol 56 No 2.

[12] Banerjee, A., Dean, H. L.,  \& Pesaran, B. (2011). "Parametric
models to relate spike train and LFP dynamics with neural information
processing." \emph{Frontiers in computational neuroscience} 6: 51-51.

[13] Theunissen, F. E. \& Miller, J.P. (1991). ``Representation of sensory
information in the cricket cercal sensory system. II. information
theoretic calculation of system accuracy and optimal tuning-curve
widths of four primary interneurons,'' \emph{J. Neurophysiol.}, vol. 66,
no. 5, pp. 1690-1703.

[14] Cortes, C., et al. "Learning curves: Asymptotic values and rate of convergence." (1994). \emph{Advances in Neural Information Processing Systems.}

\section{Appendix}

\textbf{Lemma 1. }
\emph{
Suppose $(Z_1, Z_2, \hdots, Z_k)$ are jointly multivariate normal, with 
$\E[Z_1 - Z_i]= \alpha$, 
$\Var(Z_1) = \beta$, 
$\Cov(Z_1, Z_i) = \gamma$, 
$\Var(Z_i)= \delta$, and $\Cov(Z_i, Z_j) = \epsilon$ for all $i, j = 2, \hdots,
k$, such that $\beta + \epsilon - 2\gamma > 0$.  Then, letting
\[
\mu = \frac{\E[Z_1 - Z_i]}{\sqrt{\frac{1}{2}\Var(Z_i - Z_j)}} = \frac{\alpha}{\sqrt{\delta - \epsilon}},
\]
\[
\nu^2 = \frac{\Cov(Z_1 -Z_i, Z_1 - Z_j)}{\frac{1}{2}\Var(Z_i - Z_j)} = \frac{\beta + \epsilon - 2\gamma}{\delta - \epsilon},
\]
we have
\begin{align*}
\Pr[Z_1 < \max_{i=2}^k Z_i] &= \Pr[W < M_{k-1}]
\\&= 1 - \int \frac{1}{\sqrt{2\pi\nu^2}} e^{-\frac{(w-\mu)^2}{2\nu^2}} \Phi(w)^{k-1} dw,
\end{align*}
where $W \sim N(\mu, \nu^2)$ and $M_{k-1}$ is the maximum of $k-1$
independent standard normal variates, which are independent of $W$.
}

\textbf{Proof.}
We can construct independent normal variates $G_1$, $G_2,\hdots, G_k$
such that
\[
G_1 \sim N(0, \beta + \epsilon - 2 \gamma)
\]
\[
G_i \sim N(0, \delta - \epsilon)\text{ for }i > 1
\]
such that
\[
Z_1 - Z_i = \alpha + G_1 + G_i \text{ for }i > 1.
\]
Hence
\begin{align*}
\Pr[Z_1 < \max_{i=2}^k Z_i] &= \Pr[\min_{i > 1} Z_1 - Z_i < 0].
\\&= \Pr[\min_{i=2}^{k} G_1 + G_i + \alpha < 0]
\\&= \Pr[\min_{i=2}^{k} G_i < -\alpha - G_1]
\\&= \Pr[\min_{i=2}^{k} \frac{G_i}{\sqrt{\delta - \epsilon}} < -\frac{\alpha - G_1}{\sqrt{\delta - \epsilon}}].
\end{align*}
Since $\frac{G_i}{\sqrt{\delta - \epsilon}}$ are iid standard normal variates, and since
$-\frac{\alpha - G_1}{\sqrt{\delta - \epsilon}} \sim N(\mu,\nu^2)$ for $\mu$ and $\nu^2$ given in the statement of the Lemma, the proof is completed via a straightforward computation.  $\Box$

\textbf{Theorem 1.} Let $p^{[d]}(x, y)$ be a sequence of joint densities
for $d = 1,2,\hdots$ as given above.  Further assume that
\begin{itemize}
\item[A1.] $\lim_{d \to \infty} I(X^{[d]}; Y^{[d]}) = \iota < \infty.$
\item[A2.] There exists a sequence of scaling constants $a_{ij}^{[d]}$
and $b_{ij}^{[d]}$ such that the random vector $(a_{ij}\ell_{ij}^{[d]} +
b_{ij}^{[d]})_{i, j = 1,\hdots, k}$ converges in distribution to a
multivariate normal distribution.
\item[A3.] There exists a sequence of scaling constants $a^{[d]}$, $b^{[d]}$ such that
\[
a^{[d]}u(X^{(1)}, Y^{(2)}) + b^{[d]}
\]
converges in distribution to a univariate normal distribution.
\item[A4.] For all $i \neq k$,
\[\lim_{d \to \infty}\Cov[u(X^{(i)}, Y^{(j)}), u(X^{(k)}, Y^{(j)})] = 0.\]
\end{itemize}
Then for $e_{ABE, k}$ as defined above, we have
\[
\lim_{d \to \infty} e_{ABE, k} = \pi_k(\sqrt{2 \iota})
\]
where
\[
\pi_k(c) = 1 - \int_{\mathbb{R}} \phi(z - c)  \Phi(z)^{k-1} dz
\]
where $\phi$ and $\Phi$ are the standard normal density function and
cumulative distribution function, respectively.


\textbf{Proof.}

For $i = 2,\hdots, k$, define
\[
Z_i = \log p(Y^{(1)}|X^{(i)}) - \log p(Y^{(1)}|X^{(1)}).
\]
Then, we claim that $\vec{Z} = (Z_2,\hdots, Z_k)$ converges in distribution to
\[
\vec{Z} \sim N\left(-2\iota, 
\begin{bmatrix}
4\iota & 2\iota & \cdots & 2\iota\\
2\iota & 4\iota & \cdots & 2\iota\\
\vdots & \vdots & \ddots & \vdots\\
2\iota & 2\iota & \cdots & 4\iota
\end{bmatrix}
\right).
\]
Combining the claim with the lemma (stated below this proof) yields the
desired result.

To prove the claim, it suffices to derive the limiting moments
\[\E[Z_i] \to -2\iota,\]
\[\Var[Z_i] \to 4\iota,\]
\[\Cov[Z_i, Z_j] \to 2\iota,\]
for $i \neq j$,
since then assumption A2 implies the existence of a multivariate normal
limiting distribution with the given moments.

Before deriving the limiting moments, note the following identities.
Let $X' = X^{(2)}$ and $Y = Y^{(1)}$.
\[
\E[e^{u(X', Y)}] = \int p(x) p(y) e^{u(x, y)} dx dy = \int p(x, y) dx dy = 1.
\]
Therefore, from assumption A3 and the formula for gaussian exponential
moments, we have
\[
\lim_{d \to \infty} \E[u(X', Y)]-\frac{1}{2}\Var[u(X', Y)] = 0.
\]
Let $\sigma^2 = \lim_{d \to \infty} \Var[u(X', Y)]$.
Meanwhile, by applying assumption A2,
\begin{align*}
\lim_{d \to \infty} I(X; Y) &= \lim_{d \to \infty} \int p(x, y) u(x, y) dx dy 
= \lim_{d \to \infty} \int p(x) p(y) e^{u(x, y)} u(x, y) dx dy
\\&= \lim_{d \to \infty}  \E[e^{u(X, Y')}u(X, Y')]
\\&= \int_{\mathbb{R}} e^z z \frac{1}{\sqrt{2\pi \sigma^2}} 
e^{-\frac{(z + \sigma^2/2)^2}{2\sigma^2}} \text{ (applying A2)}
\\&= \int_{\mathbb{R}} z \frac{1}{\sqrt{2\pi \sigma^2}} 
e^{-\frac{(z - \sigma^2/2)^2}{2\sigma^2}}
\\&= \frac{1}{2}\sigma^2.
\end{align*}
Therefore,
\[
\sigma^2 = 2\iota,
\]
and
\[
\lim_{d \to \infty} \E[u(X', Y)] = -\iota.
\]
Once again by applying A2, we get
\begin{align*}
\lim_{d \to \infty} \Var[u(X, Y)] 
&= \lim_{d \to \infty} \int (u(x, y) - \iota)^2 p(x, y) dx dy
\\&= \lim_{d \to \infty} \int (u(x, y) - \iota)^2 e^{u(x, y)} p(x) p(y) dx dy
\\&= \lim_{d \to \infty} \E[(u(X', Y) - \iota)^2 e^{u(X', Y)}] 
\\&= \int (z - \iota)^2 e^z \frac{1}{\sqrt{4\pi\iota}} e^{-\frac{(z+\iota)^2}{4\iota}} dz \text{ (applying A2)}
\\&= \int (z - \iota)^2 \frac{1}{\sqrt{4\pi\iota}} e^{-\frac{(z-\iota)^2}{4\iota}} dz
\\&= 2\iota.
\end{align*}

We now proceed to derive the limiting moments.
We have
\begin{align*}
\lim_{d \to \infty} \E[Z] 
&= \lim_{d \to \infty} \E[ \log p(Y|X') - \log p(Y|X)]
\\&= \lim_{d \to \infty} \E[ u(X', Y) - u(X, Y) ] = -2\iota.
\end{align*}
Also,
\begin{align*}
\lim_{d \to \infty} \Var[Z]
 &= \lim_{d \to \infty} \Var[ u(X', Y) - u(X, Y) ]
\\&= \lim_{d \to \infty} \Var[ u(X', Y)] +\Var[ u(X, Y) ]\text{ (using assumption A4) }
\\&= 4\iota,
\end{align*}
and similarly
\begin{align*}
\lim_{d \to \infty} \Cov[Z_i, Z_j]
&= \lim_{d \to \infty} \Var[ u(X, Y)]\text{ (using assumption A4) }
\\&= 2\iota.
\end{align*}
This concludes the proof. $\Box$.

\subsection{Assumptions of theorem 1}

Assumptions A1-A4 are satisfied in a variety of natural models.
One example is a multivariate Gaussian model where
\[
X \sim N(0, \Sigma_d)
\]
\[
E \sim N(0, \Sigma_e)
\]
\[
Y = X + E
\]
where $\Sigma_d$ and $\Sigma_e$ are $d \times d$ covariance matrices, and where $X$ and $E$ are independent.  Then, if $d \Sigma_d$ and $\Sigma_e$ have limiting spectra $H$ and $G$ respectively,
the joint densities $p(x, y)$ for $d = 1,\hdots, $ satisfy assumptions A1 - A4.

We can also construct a family of densities satisfying A1 - A4,
which we call an \emph{exponential family sequence model} since each joint distribution in the sequence
is a member of an exponential family.
A given exponential family sequence model is specified by choice of a base carrier function $b(x, y)$ and base sufficient statistic $t(x, y)$, with the property that carrier function factorizes as
\[
b(x, y) = b_x(x) b_y(y)
\]
for marginal densities $b_x$ and $b_y$.
Note that the dimensions of $x$ and $y$ in the base carrier function are arbitrary;
let $p$ denote the dimension of $x$ and $q$ the dimension of $y$ for the base carrier function.
Next, one specifies a sequence of scalar parameters $\kappa_1, \kappa_2,\hdots$ such that
\[
\lim_{d \to \infty} d \kappa_d = c < \infty.
\]
for some constant $c$.
For the $d$th element of the sequence, $X^{[d]}$ is a $pd$-dimensional vector,
which can be partitioned into blocks
\[
X^{[d]} = (X_1^{[d]},\hdots, X_d^{[d]})
\]
where each $X_i^{[d]}$ is $p$-dimensional.  Similarly, $Y^{[d]}$ is partitioned into $Y_i^{[d]}$ for $i = 1,\hdots, d$.
The density of $(X^{[d]}, Y^{[d]})$ is given by
\[
p^{[d]}(x^{[d]}, y^{[d]}) = Z_d^{-1} \left(\prod_{i=1}^d b(x_i^{[d]}, y_i^{[d]}) \right) 
\exp\left[\kappa_d \sum_{i=1}^d t(x_i^{[d]}, y_i^{[d]}) \right],
\]
where $Z_d$ is a normalizing constant.
Hence $p^{[d]}$ can be recognized as the member of an exponential family with carrier measure
\[
\left(\prod_{i=1}^d b(x_i^{[d]}, y_i^{[d]}) \right)
\]
and sufficient statistic
\[
\sum_{i=1}^d t(x_i^{[d]}, y_i^{[d]}).
\]

One example of such an exponential family sequence model is a
multivariate Gaussian model with limiting spectra $H = \delta_1$ and
$G = \delta_1$, but scaled so that the marginal variance of the
components of $X$ and $Y$ are equal to one.  This corresponds to a
exponential family sequence model with
\[
b_x(x) = b_y(x) = \frac{1}{\sqrt{2\pi}} e^{-x^2/2}
\]
and
\[t(x, y) = xy.\]

Another example is a multivariate logistic regression model,
given by
\[
X \sim N(0, I)
\]
\[
Y_i \sim \text{Bernoulli}(e^{\beta X_i}/(1 + e^{\beta X_i}))
\]
This corresponds to an exponential family sequence model with
\[
b_x(x) = \frac{1}{\sqrt{2\pi}} e^{-x^2/2}
\]
\[
b_y(y) = \frac{1}{2}\text{ for }y = \{0, 1\},
\]
and
\[
t(x, y) = x\delta_1(y) - x\delta_0(y).
\]
The multivariate logistic regression model (and multivariate Poisson regression model)
are especially suitable for modeling neural spike count data;
we simulate data from such a multivariate logistic regression model in section X.

\section{Additional simulation results}

Multiple-response logistic regression model
\[
X \sim N(0, I_p)
\]
\[
Y \in \{0,1\}^q
\]
\[
Y_i|X = x \sim \text{Bernoulli}(x^T B_i)
\]
where $B$ is a $p \times q$ matrix.

Multiple-response logistic regression model
\[
X \sim N(0, I_p)
\]
\[
Y \in \{0,1\}^q
\]
\[
Y_i|X = x \sim \text{Bernoulli}(x^T B_i)
\]
where $B$ is a $p \times q$ matrix.

\emph{Methods.}
\begin{itemize}
\item \text{Nonparametric}: $\hat{I}_0$ naive estimator, $\hat{I}_\alpha$ anthropic correction.
\item \text{ML-based}: $\hat{I}_{CM}$ confusion matrix, $\hat{I}_F$ Fano, $\hat{I}_{HD}$ high-dimensional method.
\end{itemize}

Sampling distribution of $\hat{I}$ for \small{$\{p = 3$, $B
= \frac{4}{\sqrt{3}} I_3$, $K = 20$, $r = 40\}$.}

True parameter $I(X; Y) = 0.800$ \emph{(dotted line.)}
\begin{center}
\includegraphics[scale = 0.5, clip = true, trim = 0 0.5in 0 0.5in]{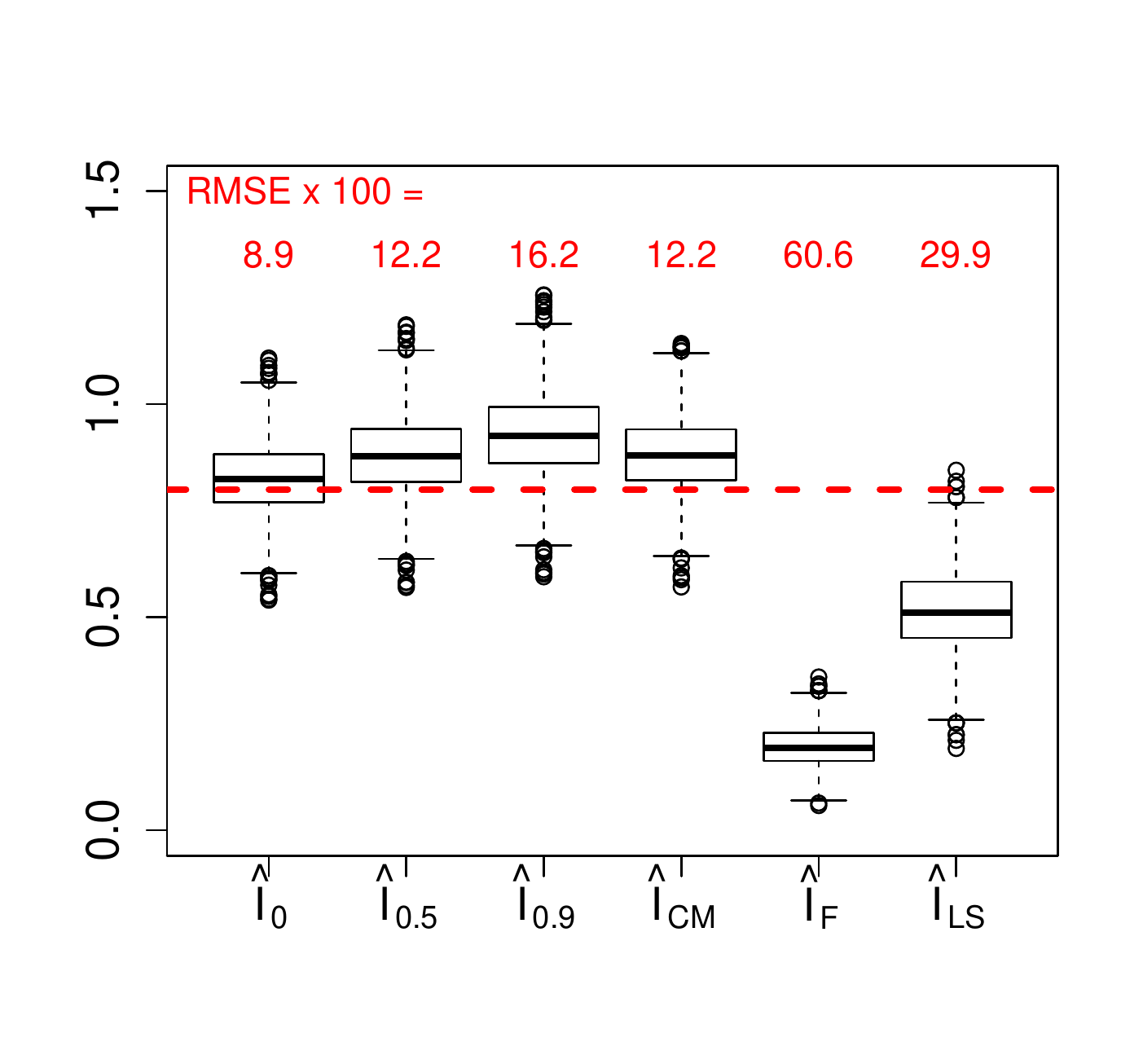}
\end{center}
Na\"{i}ve estimator performs best!  $\hat{I}_{HD}$ not effective.

Sampling distribution of $\hat{I}$ for \small{$\{p = 50$, $B = \frac{4}{\sqrt{50}} I_{50}$, $K = 20$, $r = 8000\}$.}

True parameter $I(X; Y) = 1.794$ \emph{(dashed line.)}
\begin{center}
\includegraphics[scale = 0.5, clip = true, trim = 0 0.5in 0 0.5in]{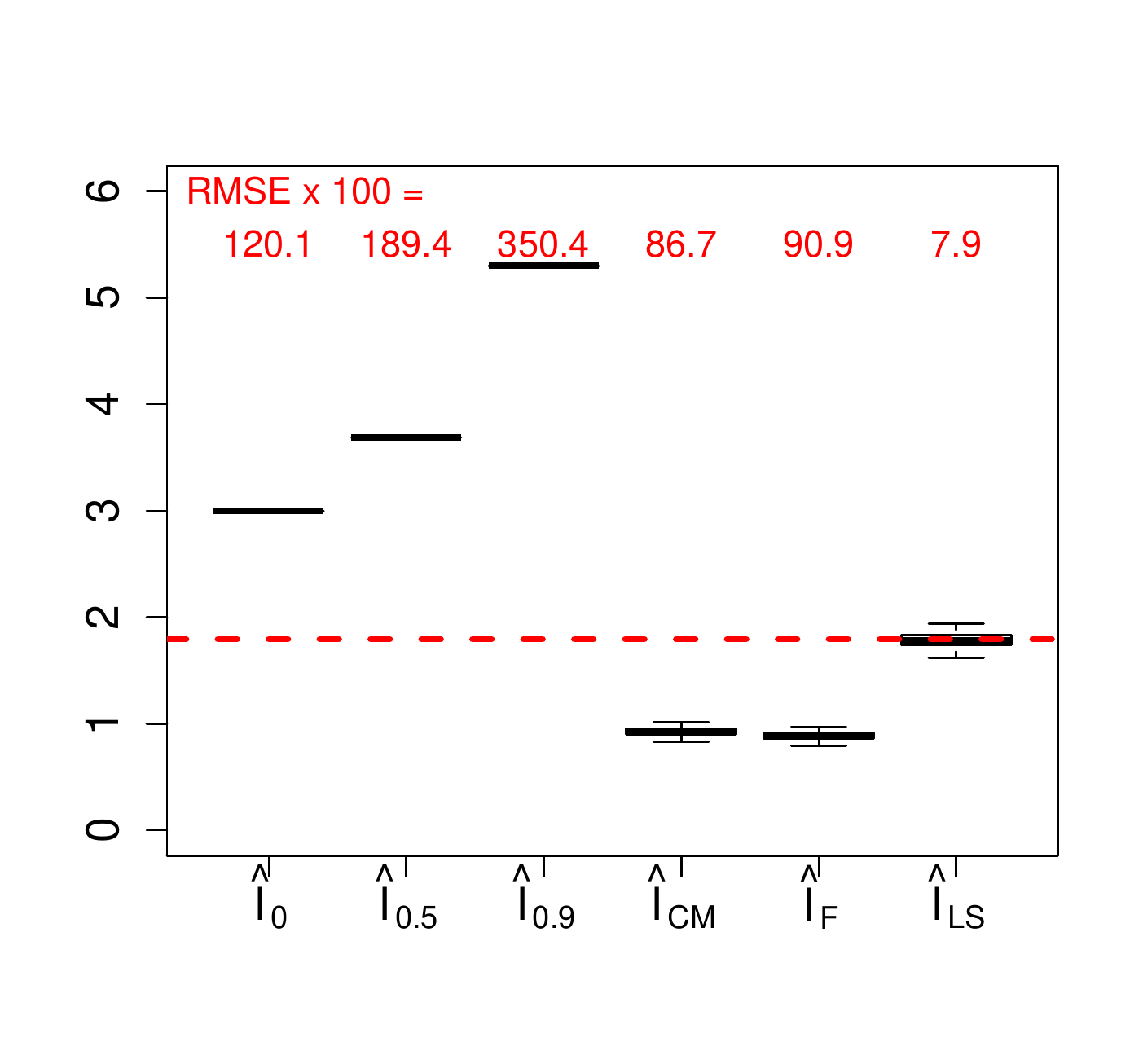}
\end{center}
Non-parametric methods extremely biased.

Estimation path of $\hat{I}_{HD}$ and $\hat{I}_\alpha$ as $n$ ranges from $10$ to $8000$.

\small{$\{p = 10$, $B = \frac{4}{\sqrt{10}} I_{10}$, $K = 20\}$.
True parameter $I(X; Y) = 1.322$ \emph{(dashed line.)}}

\begin{center}
\includegraphics[scale = 0.4]{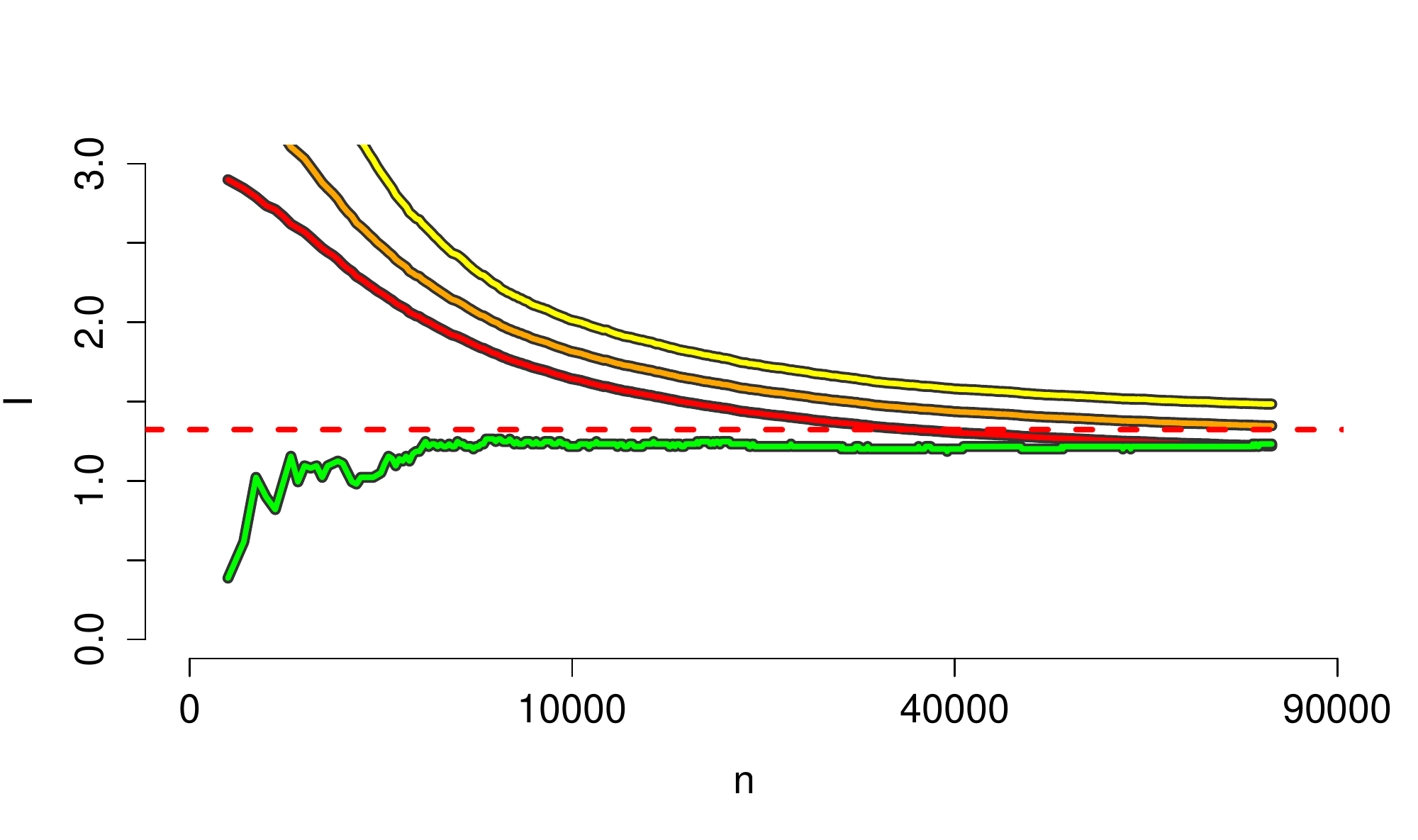}
\end{center}

\begin{center}
\textbf{Estimated $\hat{I}$ vs true $I$.} 

\includegraphics[scale = 0.5, clip=true, trim=0.4in 0.5in 0 0.5in]{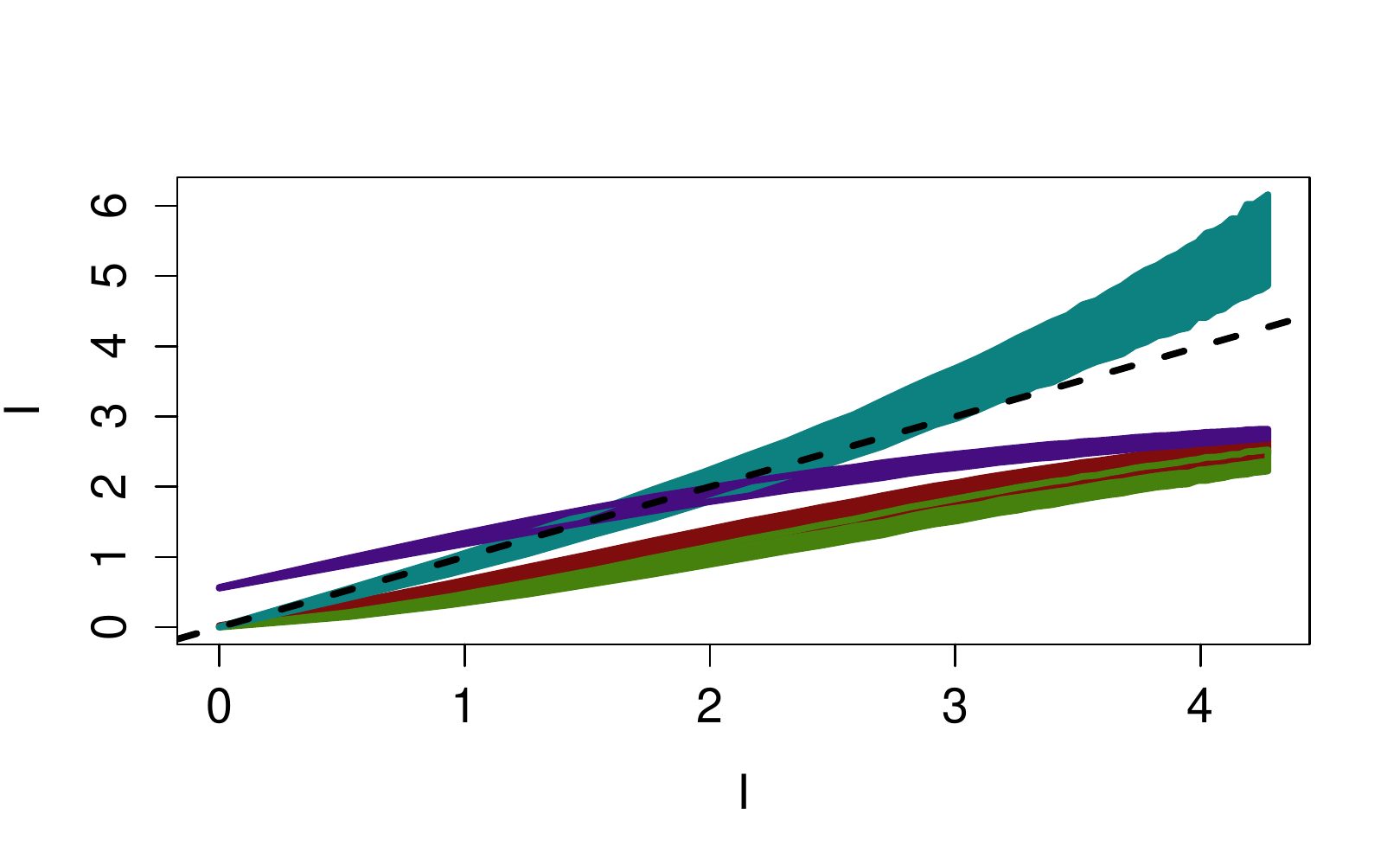}
\end{center}

Sampling distribution of $\hat{I}_{HD}$ for \small{$\{p = 10$, $B = \frac{4}{\sqrt{10}} I_{10}$, $N = 80000\}$,

and $K = \{5, 10, 15, 20, \hdots, 80\}$, $r = N/k$.}

True parameter $I(X; Y) = 1.322$ \emph{(dashed line.)}
\begin{center}
\includegraphics[scale = 0.6, clip = true, trim = 0 0.5in 0 0.5in]{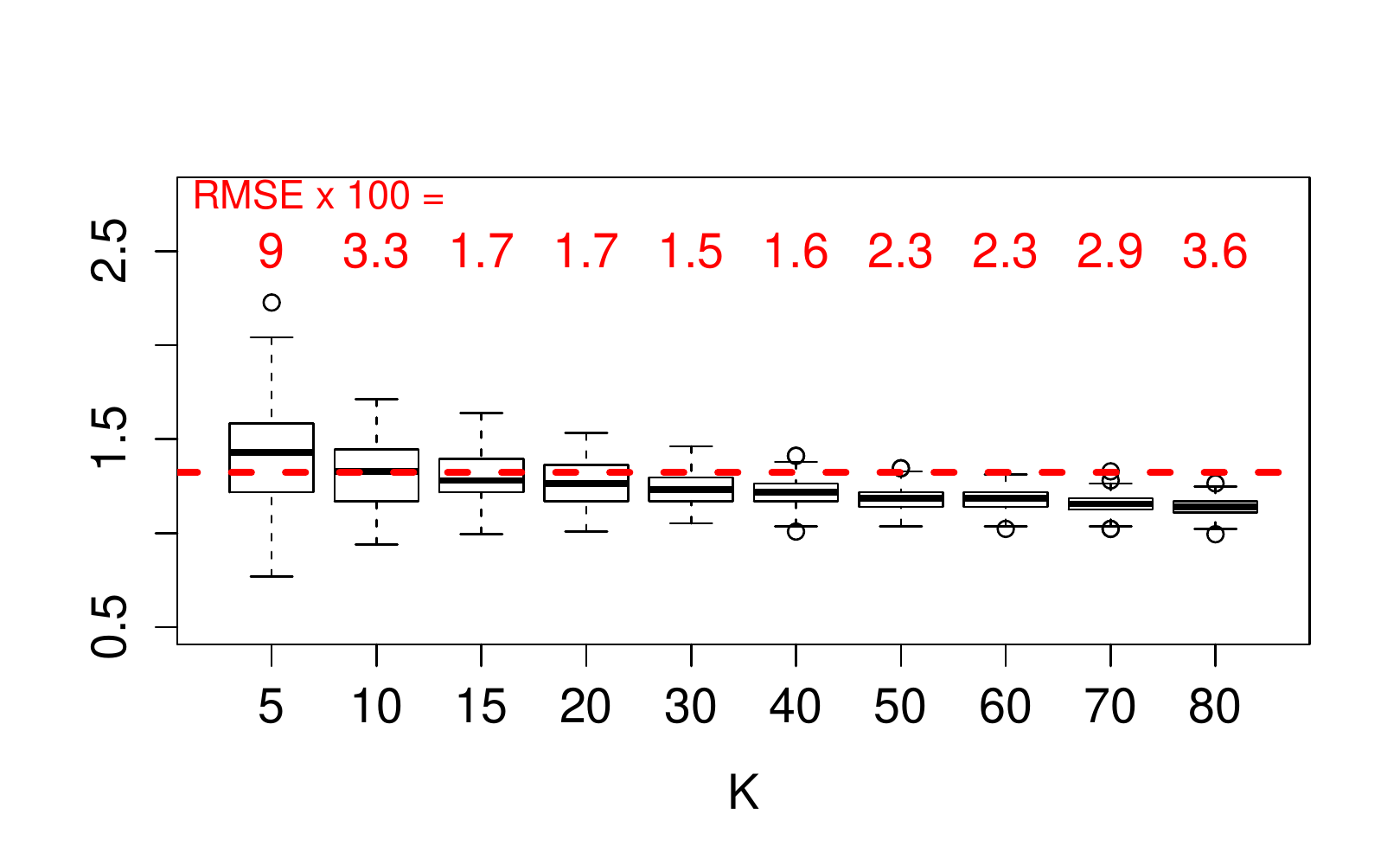}
\end{center}

Decreasing variance as $K$ increases. Bias at large and small $K$.

$p = 20$ and $q = 40$, entries of $B$ are iid $N(0, 0.025)$.

$K=20$, $r = 8000$, true $I(X; Y) = 1.86$ \emph{(dashed line.)}

\begin{center}
\textbf{Sampling distribution of $\hat{I}$.}

\includegraphics[scale = 0.6, clip = true, trim = 0 0.5in 0 0.5in]{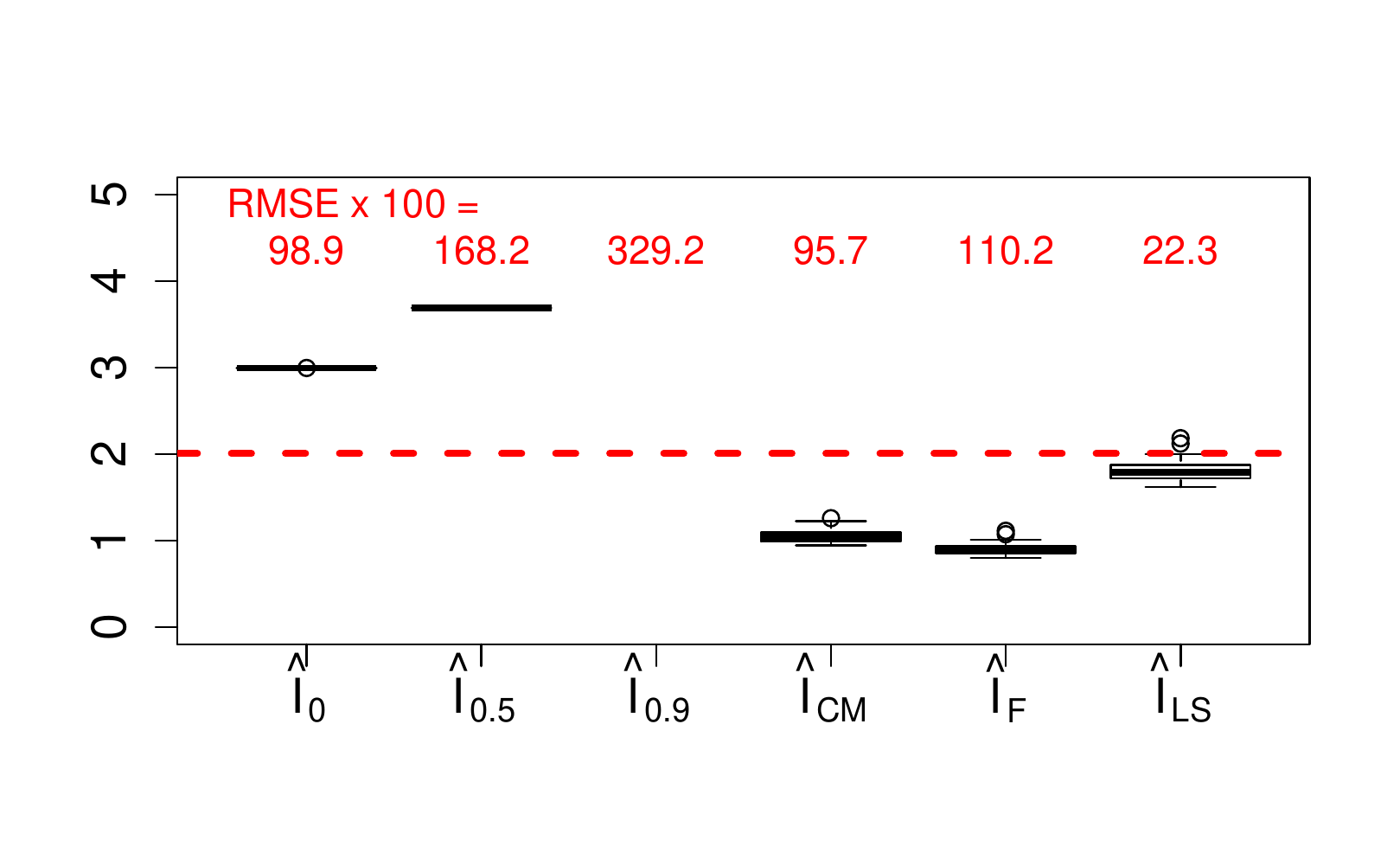}
\end{center}

\end{document}